\documentclass[10pt,twocolumn,letterpaper]{article}

\usepackage{iccv}
\usepackage{times}
\usepackage{epsfig}
\usepackage{graphicx}
\usepackage{amsmath}
\usepackage{amssymb}
\usepackage{multirow}
\usepackage{mathtools}

\usepackage[pagebackref=true,breaklinks=true,letterpaper=true,colorlinks,bookmarks=false]{hyperref}

\iccvfinalcopy 

\def\iccvPaperID{9561} 

\ificcvfinal\fi

\renewcommand{\epsilon}{\varepsilon}
\renewcommand{\phi}{\varphi}

\DeclareMathOperator{\argmax}{\arg\,\max}

\newcommand{\calD}{\mathcal{D}}
\newcommand{\calL}{\mathcal{L}}

\newcommand{\calA}{\mathcal{A}}
\newcommand{\calT}{\mathcal{T}}
\newcommand{\calP}{\mathcal{P}}
\newcommand{\calW}{\mathcal{W}}

\newcommand{\bbR}{\mathbb{R}}
\newcommand{\bbI}{\mathbb{I}}

\newcommand{\bp}{\mathbf{p}}

\newcommand{\by}{\mathbf{y}}

\newcommand{\feat}{\mathbf{f}}

\newcommand{\vid}{\ensuremath{\text{V}}}
\newcommand{\enc}{\ensuremath{\mathbf{\Phi}}}
\newcommand{\dec}{\ensuremath{\mathbf{\Psi}}}
\newcommand{\bottle}{\ensuremath{\mathbf{\Gamma}}}
\newcommand{\finp}{\ensuremath{\feat^{in}}}
\newcommand{\tenc}{\ensuremath{T^{en}}}
\newcommand{\denc}{\ensuremath{d^{en}}}
\newcommand{\fenc}{\ensuremath{\feat^{en}}}

\newcommand{\acc}{\ensuremath{\text{Acc}}}

\newcommand{\PP}{\ensuremath{\mathbb{P}}}
\newcommand{\CP}[2]{\PP\!\left(\left. #1 \, \right| #2 \right)}

\newcommand{\EE}{\ensuremath{\mathbb{E}}}

\newcommand{\BK}[1]{ {\left( #1 \right)} }
\newcommand{\sqBK}[1]{ {\left[ #1 \right]} }

\newcommand{\norm}[1]{\left\Vert #1 \right\Vert}
\DeclarePairedDelimiter\ceil{\lceil}{\rceil}
\DeclarePairedDelimiter\floor{\lfloor}{\rfloor}

\newcommand{\modelname}{C2F-TCN}

\begin{document}

\title{Coarse to Fine Multi-Resolution Temporal Convolutional Network}
\author{
Dipika Singhania$^{*}$, Rahul Rahaman$^{}$\thanks{ indicates equal contribution. }, Angela Yao$^{}$\\
${}$National University of Singapore\\
{\tt\small dipika16@comp.nus.edu.sg, rahul.rahaman@u.nus.edu, ayao@comp.nus.edu.sg}
}


\maketitle
\ificcvfinal\thispagestyle{empty}\fi

\begin{abstract}
    Temporal convolutional networks (TCNs) are a commonly used architecture for temporal video segmentation.  TCNs however, tend to suffer from over-segmentation errors and require additional refinement modules to ensure smoothness and temporal coherency. In this work, we propose a novel temporal encoder-decoder to tackle the problem of sequence fragmentation.  In particular, the decoder follows a coarse-to-fine structure with an implicit ensemble of multiple temporal resolutions.  The ensembling produces smoother segmentations that are more accurate and better-calibrated, bypassing the need for additional refinement modules. In addition, we enhance our training with a multi-resolution feature-augmentation strategy to promote robustness to varying temporal resolutions.  Finally, to support our architecture and encourage further sequence coherency, we propose an action loss that penalizes misclassifications at the video level. Experiments show that our stand-alone architecture, together with our novel feature-augmentation strategy and new loss, outperforms the state-of-the-art on three temporal video segmentation benchmarks. 
\end{abstract}

\section{Introduction}\label{sec:introduction}

We tackle the problem of temporal video segmentation and classification for untrimmed videos of complex activities. A complex activity is usually goal-oriented, \eg \emph{`frying eggs'} and composed of multiple steps or sub-actions in some sequence, \eg \emph{`Pour Oil'}, \emph{`Crack Egg'}, ... 
\emph{`Put to Plate'} over time.  The standard framework for temporal segmentation is MS-TCN~\cite{li2020ms, farha2019ms}. MS-TCN uses
temporal convolutions with progressively larger dilations to maintain a constant temporal resolution throughout the feedforward architecture. Multiple works further augment the MS-TCN model \cite{wang2020boundary, wang2020gated} with additional model training or post-processing smoothing. 

We posit that maintaining a constant temporal resolution is suboptimal for handling video sequences.  In this work, we propose an encoder-decoder instead: the encoder reduces the temporal resolution to some bottleneck feature before a symmetric decoder gradually recovers the sequence back to the original temporal resolution. Such a shrink-then-stretch strategy is in line with many other works in vision for image segmentation~\cite{ronneberger2015u, dac-spp-unet}, depth and flow estimation~\cite{tu2019real} and landmark detection~\cite{newell2016stacked, landmarkdetect-yang2017stacked}.  

In fact, encoder-decoders have also been used for video sequence understanding in the past~\cite{TED-lea2017temporal, TEDresi-lei2018temporal, TED-ding2018weakly}, though their performance is not strong. 
We believe that possible causes include the simple bottleneck and decoder design. Furthermore, the decoder must be carefully designed for action segmentation because of the significant variation in the sub-action lengths.  

To handle different sub-actions lengths, we propose a novel \textit{``coarse-to-fine ensemble''} of decoder output layers. Our ensemble is not only more accurate, but also has a smoothing effect and yields more accurately calibrated sequences. Incorporation of \textit{``coarse-to-fine ensemble''} of decoder outputs is a key novelty of our work in comparison to previous temporal Encoder-Decoder architectures~\cite{TED-lea2017temporal, TEDresi-lei2018temporal, TED-ding2018weakly}. To support our proposed architecture, we incorporate the following two simple yet effective novelties. 

\textbf{Video-level loss:}
Firstly, we propose a video-level ``Action Loss''.  Currently, temporal convolutional frameworks for videos are all trained with frame-level losses; however, these do not adequately penalize sequence-level miss-classification. To augment the frame-level losses, we introduce a novel video level ``Action Loss'' to penalize sub-actions not associated with the complex activity label. Such a loss is highly advantageous in mitigating the effects of over-segmentation and prevent fragmented sequence segmentation. 

\textbf{Multi-resolution feature augmentation:} Although augmentations are quite common in deep learning, as per our knowledge, none of the previous work considers data augmentation for sequence segmentation, likely due to the following reasons. A direct translation of augmentation techniques from images would require perturbations to the video frames~\cite{videoaug-han2019video, videoaug-wang2016temporal}. This would not only be computationally expensive but suffer domain gap problems since the standard is to use pre-computed features from pre-trained networks.
As an alternative, we consider augmentation at a feature level~\cite{featureaug-dong2019feature} specifically for video sequences. 
Specifically, we propose \textit{``multi-resolution feature-augmentation''} strategy, which augments video sequences by considering features of different sampling rates.  This adds robustness to the model and allows the network to learn to handle sub-sampled resolutions of video. Previous works~\cite{farha2019ms} found that using lower frame-rates drops the accuracy,  motivating the need for a full (temporal) resolution framework which is computationally expensive. We can derive higher accuracy with frame rates much lower than that of the original video through our augmentation strategy.

\textbf{Uncertainty quantification in Action Segmentation:} Due to the direct application of video action segmentation in real-life human activities, models with over-confidently wrong predictions can lead to disastrous consequences. Thus, it is highly crucial for the models to have good prediction uncertainty. Traditional neural networks are known to be over-confident in their predictions \cite{Guo_calibration_2017}. Among all the available solutions that combat over-confidence, probability ensembles are one of the most effective \cite{DeepEnsemble, lee2015m} especially in the presence of abundant training set \cite{rahaman2020uncertainty}. We thus highlight the calibration performance of our proposed method in this work.

To summarize, our main contributions are 

    \textbf{(1)} We design an novel temporal Encoder-Decoder architecture \modelname{} for temporal segmentation. This utilizes the coarse-to-fine resolutions of decoder outputs to give more calibrated, less fragmented, and accurate outputs.
    
    \textbf{(2)} We propose a multi-resolution temporal feature-level augmentation strategy that is computationally more efficient and gives a significant improvement in accuracy.
    
    \textbf{(3)} We propose an \emph{``Action Loss''} which penalizes misclassifications at the video level and enhances our frame-level predictions. This global video-level loss serves as an auxiliary loss that complements the frame-level loss without any additional network structure.
    
    \textbf{(4)} We adapt our temporal segmentation model \modelname{} to recognize the minutes-long video and achieves SOTA performance.
    
\section{Related Work}
\begin{figure*}
\begin{center}
\includegraphics[trim={0 9.2cm 0 0},width=1.0\linewidth]{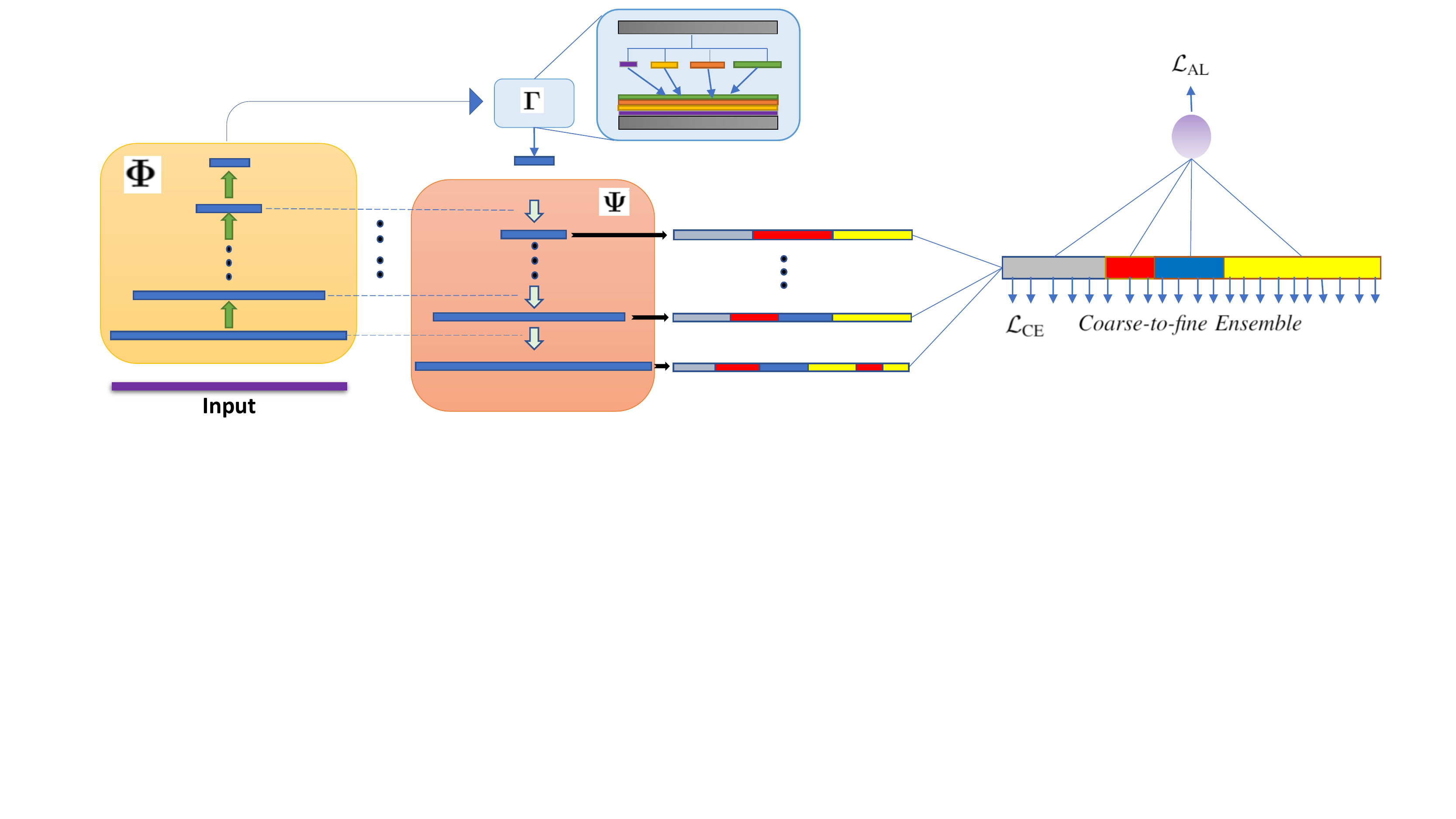}
\end{center}
\caption{\textbf{Our segmentation architecture:} Depiction of the architecture of our model $(\enc:\bottle:\dec)$. We utilize our multi-resolution features to produce \textit{Coarse-to-fine Ensemble} predictions.}

\label{fig:main_architecture}
\end{figure*}
\textbf{Action Recognition.} The task of recognition requires classifying the video as whole whereas task of segmentation requires classifying each frame of the video. For short trimmed videos($2-10$ sec) architectures for action recognition include two-stream 2D CNNs~\cite{trimmed2d-feichtenhofer2016convolutional, trimmed2d-simonyan2014two}, CNNs together with LSTMs~\cite{trimmed2dlstm-donahue2015long, trimmed2dlstm-ullah2017action}, 3D CNNs~\cite{trimmed3d-ji20123d, trimmed3d-tran2015learning}.  More recent architectures include a two-stream 3D-CNN (I3D)~\cite{carreira2017quo}
, non-local blocks~\cite{trimmednon-wang2018non}, SlowFast network \cite{longvideosrecog-feichtenhofer2019slowfast} and the Temporal Shift Module \cite{lin2019tsm}.

For longer, untrimmed videos, the computational cost of directly training such deep $3D$ convolutional architectures is computationally very expensive and data deficit. Most approaches resort to extracting snippet-level features based on the previously-mentioned architectures and add additional dedicated sequence-level processing for either recognition or segmentation.
Examples include TCN networks\cite{TED-ding2018weakly, TED-lea2017temporal, TEDresi-lei2018temporal, li2020ms, farha2019ms}, RNNs \cite{prevsegmentrnn-kuehne2018hybrid, prevsegmentrnn-singh2016multi, prevsegmentrnn-perrett2017recurrent}, graphs~\cite{highlevel-hussein2019videograph}, dedicated architectures like the temporal inception network~\cite{highlevel-hussein2019timeception} or attention modelling ~\cite{highlevel-hussein2020pic, sener2020temporal}. Our work falls into this category where we model temporal relationships on top of snippet level I3D features. 

For only long video recognition, some use temporal convolutions\cite{highlevel-hussein2019timeception}, attention \cite{highlevel-hussein2020pic, highlevel-hussein2019videograph} or graph modelling \cite{highlevel-hussein2019videograph}; others use aggregating models likes ActionVLAD \cite{highlevel-girdhar2017actionvlad} and some generating framework for recognition \cite{kuehne2014language}. Different from these we adapt our segmentation model to be used for temporal action recognition task as it captures both global relationships and local fine-grained information for recognizing the video action.

\textbf{Action Segmentation.}
Initially, models for temporal video segmentation were based on statistical 
length modelling~\cite{prevsegment-statis-richard2016temporal}, RNNs \cite{prevsegmentrnn-perrett2017recurrent, prevsegmentrnn-singh2016multi, prevsegmentrnn-perrett2017recurrent}, and RNNs combined with Hidden Markov models~\cite{prevsegmentrnn-kuehne2018hybrid}. The recursive nature of RNNs make them slow and difficult to train, especially for long sequences. 
More recent architectures treat the entire video sequence as a whole and model coarse units within the sequence either via non-local attention~\cite{sener2020temporal} or graph convolutional networks~\cite{huang2020improving}.
However, these approaches cannot account for the fine-grained temporal changes in videos, required for segmentation.

Several works have shown the effectiveness of temporal convolutional networks (TCN) ~\cite{TED-ding2018weakly, TED-lea2017temporal, TEDresi-lei2018temporal, farha2019ms, li2020ms}.  MS-TCN~\cite{farha2019ms, li2020ms} stacks a temporal convolutions with dilation, without max-pool to capture the relationships in long videos. Encoder-decoder TCN's ~\cite{TED-ding2018weakly, TED-lea2017temporal, TEDresi-lei2018temporal} uses gradual reduce and expanded resolution convolutions.
In line with Encoder-Decoder TCN's, we design our model architecture \modelname{}. However, we are first to explore the coarse-to-fine decoder layer's ensemble. Segmentation with TCN still suffers from over-fragmentation errors, which are handled with refinement layers \cite{farha2019ms, li2020ms}, LSTMs on top of refinement layers \cite{wang2020gated}, or separate boundary detection model with post-processing smoothing \cite{wang2020boundary, ishikawa2021alleviating}. Different from these, we handle over-fragmentation with our implicit model's components ensembling which does not require any additional network structure.

\section{Methodology}\label{sec:method}

The aim of video action segmentation is to assign a label to each frames of a video. More formally, it deals with video $\text{V} \in \bbR^{T \times H\times W}$, where for each temporal location $t \in \calT := \{1,...,T\}$, frame $\vid_t \in \bbR^{H\times W}$ is an image of dimension $H\times W$. With a pre-defined set of $C$ action classes $\calA$ := $\{1,...,C\}$, the task of action segmentation is to find a mapping $\hat{y}: \calT \to \calA$ that maps each frame $\vid_t$ to an action label $\hat{y}_t \in \calA$.
Our method is supervised; the model $M$ takes a video $\vid$ as input and produces predictions $M(\vid) = \{\bp_t\}^{T}_{t=1}$ where each $\bp_t \in \bbR^C$ is a probability vector of dimension $C$. The predicted label for each $t$ is then obtained by $\hat{y}_t = \argmax \;\bp_t$ and the corresponding probability by $\hat{p}_t = \max \; \bp_t$, with the $\max$ and $\argmax$ over all possible actions in $\calA$. To overcome the computational challenge of training an end-to-end model, the standard practice is to use pre-trained frame level features. We work with such pre-trained feature representation of every frame $t$ denoted by $\feat_t \in \bbR^d$. Furthermore, instead of using the features at full temporal length $T$, we down-sample $\feat$ to obtain coarser features $\feat^{in}$ of temporal dimension $T^{in}$, which we use as input to our model. We discuss the details of our down-sampling strategy in subsection \ref{subsec:augmentation}.  However, it should be noted that for each video during inference, we always up-sample our predictions $\{\hat{y}_t: t \le T_{in}\}$ to the original temporal dimension $T$, in order to compare with original ground truth $\{y_t\}^T_{t=1}$. 

\subsection{Model Architecture:}\label{subsec:model_architec}
Our model's architecture can be separated into three components, \ie $M := (\enc:\bottle:\dec)$, consisting of the encoder \enc{}, bottleneck \bottle{} and decoder \dec{}. Below we discuss each component of the architecture in more details.

\textbf{Encoder network \enc:} The input to the encoder network is the down-sampled frame-level features $\finp \in \bbR^{T^{in}\times d}$, where $d$ denotes channel dimension of input feature. The encoder network consists of a 1-D convolution unit $\enc^{(0)}$ and six sequential encoder blocks $\{\enc^{(i)}\!:\! i\!\le\!6\}$. At the beginning, $\enc^{(0)}$ projects \finp{} to feature of dimension $T^{in} \times d_0$;
for $i\!\ge\!1$, the outputs of $\enc^{(i)}$ are $\bbR^{T_i \times d_i}$, where $T_i$ and $d_i$ are the temporal and feature dimensions of each block $i$ respectively. 
Each encoder consists of a convolutional block and a max-pooling unit that halves its input's temporal dimension.

\textbf{Bottleneck network \bottle:} 
To ensure flexibility in the input video resolution and to facilitate temporal augmentations, we introduce a pyramid pooling \cite{spp_original}
along the temporal dimension. Pyramid pooling has been used in the past to handle multi-resolution inputs for segmentation, detection, and recognition task in both images and videos \cite{spatial-pp, temporal-pp, deeplab, deeplabv3, dac-spp-unet, zheng2019spatial}.

The input to the bottleneck \bottle{} is the output \fenc{} of the last encoder layer $\enc^{(6)}$.  We create four redundant representations of \fenc{} of varying temporal resolutions via 
four parallel max-pooling units of varying kernel sizes $\{w^{\gamma}_i: i\le 4\}$. The max-pooling reduces \fenc{} 
to a smaller temporal dimension $\floor*{\frac{\tenc}{w^{\gamma}_i}}$; each feature is then collapsed by a shared 1D convolution of kernel size 1 to a single latent dimension while the temporal dimension is kept fixed. 
These four redundant representations are then up-sampled with linear interpolation back to the original temporal dimension \tenc{}. Along with the features \fenc{}, the four features of dimension $\tenc{} \times 1$ are concatenated along latent dimension to produce a bottleneck output of $\mathbb{R}^{\tenc{} \times (4+\denc)}$. 
Owing to the shared nature of the single 1-D convolution unit, the bottleneck network contains marginal parameters.

\textbf{Decoder network \dec:} Structurally, the decoder network is symmetric to the encoder; it has six decoder layers $\{\dec^{(i)}: i\!\le\!6\}$, each containing an up-sampling unit and the same convolution block as the encoder (see Fig. \ref{fig:main_architecture}). 
For each $i\!\ge\!1$, the up-sampling unit linearly interpolates or \emph{stretches} inputs to an output of twice the temporal length.
This is then concatenated with $\enc^{(6-i)}$, \ie the output of the $(6-i)^{th}$ encoder block via a skip connection. 
The output of the $i^{th}$ decoder block thus has temporal dimension same as $T_{6-i}$ and latent dimension $128$. 
The skip-connection ensures that both global information (from the decoder) and local information (from the encoder).  The last decoder layer $\dec^{(6)}$, with a skip connection from $\enc^{(0)}$ generates output of $\mathbb{R} ^{T^{in}\times 128}$. 

\subsection{Coarse-to-Fine Ensemble (C2F Ensemble):}\label{subsec:multilayer_ensemble}
Rather than taking the last decoder layer as the output, we propose to ensemble the results from several decoder layers. For each $i\!\ge\!1$, we project the output of the $i^{th}$ decoder block $\dec^{(i)}$ to 
$C$ dimensions, \ie the number of actions and follow with a 
softmax operation to produces probability vectors $\bp^{(i)}$ which 
get up-sampled with linear interpolation to the input temporal dimension $T^{in}$ and ensembled. Thus, for any $1\!\le\!t\!\le T^{in}$, the ensembled
prediction ${\bp}^{ens}_t \in \bbR^{C}$ takes the form -
\begin{gather}\label{eqn:ensemble}
    {\bp}^{ens}_t = \sum_{i} \alpha_i \cdot \text{Up}\sqBK{\bp^{(i)}, t}, \quad \sum_i \alpha_i = 1, \; \alpha_i > 0
\end{gather}
where $\alpha_i$ are the ensemble weights of the $i^{th}$ decoder layer, and $\text{Up}\BK{\cdot, t}$ is a function that returns the interpolated vector at time $t$. The sum in Eq.~\ref{eqn:ensemble} is performed action class-wise, with the final predicted action label as
\begin{equation}
 \hat{y}_t = \underset{k \in \calA}{\argmax} \, {\bp}^{ens}_t.
 \end{equation}

The ensembled predictions ${\bp}^{ens}$ is used during both training and inference; we refer to it as a \textit{coarse-to-fine ensemble} (C2F ensemble). Our rationale for using an ensemble is twofold. First, the earlier decoder layers are less susceptible to fragmentation errors by being coarser in their temporal resolution. Incorporating these outputs helps to mitigate over-segmentation. Such multi-resolution ensemble is made possible due to the \textit{shrink-then-stretch} policy of encoder-decoder architecture. In contrast, architectures such as MS-TCN keep the temporal dimension unchanged, requiring additional refinement stages to correct over-segmentation. 
Secondly, standard network outputs tend to be over-confident in their predictions~\cite{Guo_calibration_2017} which we discuss later in section \ref{subsec:uncertainty_quantification}.  Probability ensembles are an effective way~\cite{DeepEnsemble, lee2015m} to reduce overconfidence, especially with sufficient training data~\cite{rahaman2020uncertainty}. 

\begin{figure*}
\begin{center}
\includegraphics[trim={0 10cm 0 0},width=0.9\linewidth]{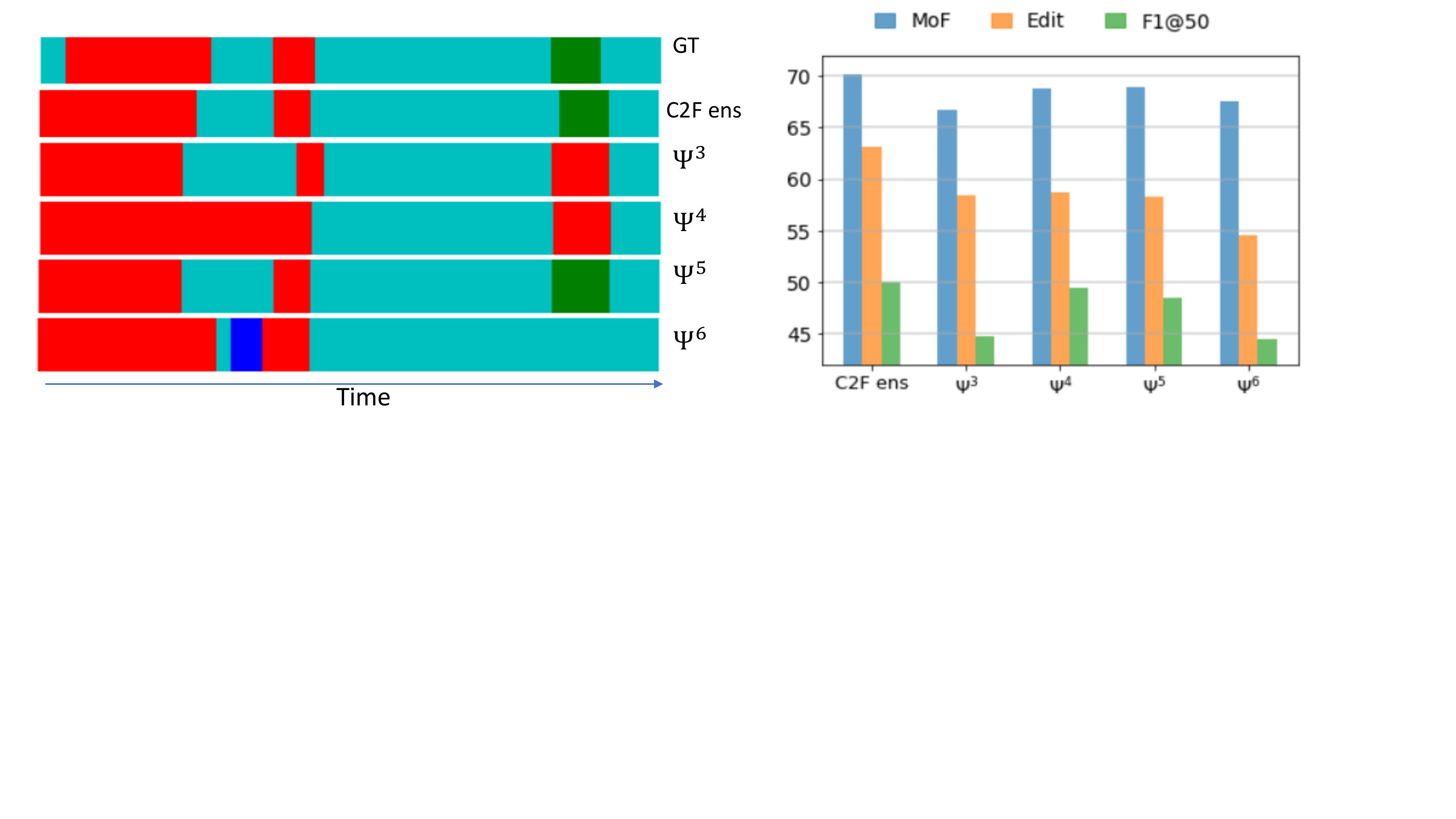}
\end{center}
\caption{\textbf{Performance of different decoder layers}: The left plot shows qualitative example of our model's video segmentation result, where each color denotes an sub-action. We see C2F ensemble(C2F ens) matches best to the ground truth(GT) than other layers. Additionally, over-fragmentation (blue) patch removed from the last decoder layer ($\Psi^6$). The right bar chart shows quantitative overall performance of different layers and C2F ensemble.}
\label{fig:layer-analysis}
\end{figure*}

\subsection{Multi-Resolution Feature Augmentation:}\label{subsec:augmentation}
As discussed earlier, we down-sample the pre-trained feature representations $\feat$ to obtain the input feature vector $\feat^{in}_t \in \bbR^{d}$ and ground truth $y^{in}_t \in \calA$. Instead of standard subsampling, we use max-pooling, which has been found to be beneficial for representing video segments in \cite{sener2020temporal}. Instead of max-pooling with a fixed window size, we consider multiple resolution of the features by varying the window. At time $t$, for some temporal window $w > 0$, we max-pool along the temporal dimension within a temporal window of $w$: 
\begin{equation}\label{eqn:downsample}
    \feat^{w}_t = \max_{\tau \in \left[wt, wt+w\right)} \feat_{\tau},
\end{equation}

\noindent while taking the ground truth action that is most frequent in the window $\left[wt, wt+w\right)$ as the corresponding label
\begin{equation}
    y^{w}_t = \underset{k \in \calA}{\argmax} \sum_{\tau=wt}^{wt+w} \bbI\sqBK{y_{\tau}=k}
\end{equation}

\noindent The pooled features $\feat^{w} := \{\feat^{w}_t\}$ and ground truth $\by^{w} := \{y^w_t\}$ both have a temporal dimension of $T^w := \ceil*{\frac{T}{w}}$. By using a varying set of windows $\calW \subset \mathbb{N}$, a corresponding set of feature-ground truth pair $\calD_{aug} := \{(\feat^w, \by^w): w \in \calW\}$ can thus be obtained. 

By equipping $\calW$, and by extension $\calD_{aug}$, with a probability distribution $\pi$, we formulate a stochastic augmentation strategy. We work with a specific class of probability distribution $\Pi:=\{\pi\BK{\cdot\;; w_0}: w_0 \in \mathbb{N}\}$ parameterized by a ``base window'' $w_0$. For a base window, we define an upper and lower bound $w_{max} := 2w_0$ and $w_{min} := \floor*{\frac{w_0}{2}}$, and $r:=w_{max} - w_{min}$. Then we define distribution $\pi$ as
\begin{gather*}
    \pi\BK{w; w_0} = \left\{\begin{array}{ll}
         \pi_0 & : \,\,w=w_0 \\
         (1 - \pi_0) /r & : \,\, \text{otherwise}
    \end{array}\right.
\end{gather*}
where $0\le \pi_0 \le 1$ is the probability of sampling the base window. We assign zero probability to any window size outside the bounds. In our experiments we found $\pi_0 = \frac{1}{2}$ to be most effective.

Apart from the obvious advantage of the increased effective size of the training data, this augmentation strategy encourages model robustness with respect to a wide range of temporal resolutions.  At test time, we can also combine predictions for different temporal windows.  

Before combining, all the predictions are interpolated to the original temporal dimension $T$ of the test example.  We compute the expectation of the prediction probabilities under the distribution $\pi$. Formally, for any $t \le T$ and $k \in \calA$ our final predictive likelihood is 
\begin{gather}\label{eqn:augmented_predict}
    \PP\BK{\hat{y}_t = k} = \EE_{w\sim \pi} \Big[\CP{\hat{y}_t = k}{\feat^{w}}\Big]
\end{gather}
where $\CP{\hat{y}_t = k}{\feat^{w}}$ is the conditional probability of the frame $t$ being labelled with action class $k$ given input features $\feat^{w}$ max-pooled with window $w$. 

\subsection{Training Losses:}\label{sec:losses}
During our training procedure, we make use of three different losses. 
The first, $\calL_{\text{CE}}$, is the standard frame-level cross-entropy for action classification:
\begin{equation}\label{eqn:cross_entropy}
    \mathcal{L}_{\text{CE}} = - \frac{1}{T} \sum_t \sum_{k\in \calA} \bbI\sqBK{y_t = k} \log \PP\sqBK{\hat{y}_t = k}
\end{equation}
where $y_t, \hat{y}_t$ are the ground truth and the predicted label respectively. The second is the transition loss used in~\cite{farha2019ms, li2020ms} $\mathcal{L}_{\text{TR}}$:
\begin{equation}\label{eqn:transfer_loss}
    \begin{split}
        \mathcal{L}_{\text{TR}}=\frac{1}{T} \sum_{t} \norm{\min\BK{\boldsymbol{\epsilon}_{t}, \epsilon_{max}}}^2 \\
        \boldsymbol{\epsilon}_{t} :=\left|\log \bp^{ens}_t - \log \bp^{ens}_{t-1}\right|.
    \end{split}
\end{equation}
to encourage neighbouring frames to have the same action label. Here,  $\bp^{ens}_t \in \bbR^C_+$ is our multi-resolution probability outputs, and the $\epsilon_{max} \in \bbR_+$ is the cutoff 
used to clip the inter-temporal absolute difference of log-probabilities. All the vector operations in equation \ref{eqn:transfer_loss} are performed element-wise.

\subsubsection{Video-Level Action Loss $\calL_{\text{AL}}$}\label{subsec:action_loss} 
The complex activity itself serves as a very strong cue for the actions present depending on the complex activity.  For example, \emph{`crack egg'} should not appear in \emph{`making coffee'}, but the standard frame-level cross-entropy would penalizes it the same as other wrong but feasible actions. 
To stronger enforce the relationships of complex activities to the actions, we propose a novel video-level auxiliary loss.
For a video $\vid$ we define $\calA_{\vid} \subset \calA$ to be the set of unique ground truth actions present in the video. Let $\delta^{pres}_k := \bbI\sqBK{k \in \calA_{\vid}}$ be the indicator whether the action $k$ is present in video $\vid$ and $\pi^{pres}_k := \max_{t} \PP\sqBK{\hat{y}_t = k}$ be the maximum action probability assigned to class $k$ across the whole video. We define our {Video-level Action Loss} as
\begin{equation}
    \begin{split}
        \calL_{\text{AL}} = -&\sum_{k\in \calA} \delta^{pres}_k \cdot \log\pi^{pres}_k \,\, \\
        & - \sum_{k\in \calA} (1 - \delta^{pres}_k) \cdot \log(1 - \pi^{pres}_k).
    \end{split}
\end{equation}
This loss ensures that we maximize the $\pi^{pres}_k$ for all actions present in a video.  More importantly, though, it allows us to minimize $\pi^{pres}_k$ for any action $k$ not present in the entire video, thereby limiting misclassifications by actions not related to the complex activity.

The three loss terms can be summed into a joint loss $\calL$:
\begin{equation}
    \calL = \calL_{\text{AL}} + \mathcal{L}_{\text{CE}} + \lambda_{\text{TR}} \mathcal{L}_{\text{TR}}
\end{equation}
where $\lambda_{\text{TR}}\!=\!0.15$ as suggested by~\cite{li2020ms, farha2019ms}.

\subsection{Complex Activity Recognition:} 
Our framework can be adopted easily from video segmentation to classify the overall complex activity.  We use the encoder-decoder architecture as described in Section \ref{subsec:model_architec} to obtain output $\bp_t$ and then max-pool over time before applying a two-layer MLP followed by a softmax:
\begin{equation}
             \bp_{\text{V}} = \text{MLP}\big[\max_t (\log {\bp}_t)\big],
\end{equation}

\noindent where $\bp_{\text{V}} \in \bbR^{K}_{+}$ is the probability vector for the $K$ complex activities.  Intuitively, max-pooling along the temporal dimension retains the important information over time and is is invariant to permutation of action orders within the video.
Similar to other works~\cite{highlevel-hussein2019timeception, highlevel-hussein2020pic}, 
we do not use any frame wise sub-action labels.  Instead, we train our segmentation network separately with the following loss
\begin{equation}
 \mathcal{L_{V}} = - \sum_{k=1}^K \bbI\sqBK{y_{\text{V}} = k} \log \PP\sqBK{\hat{y}_{\text{V}} = k}.
\end{equation}

\noindent where $y_{\text{V}}, \hat{y}_{\text{V}}$ are the ground truth and the predicted complex activity of video sequence $\text{V}$.

\subsection{Calibration:}\label{subsec:calibration_notation}
The calibration of a prediction is a measurement of over/under-confidence. Earlier we defined maximum probability prediction of a frame $t$ to be $\hat{p}_t := \max \bp_t$. In calibration literature, $\hat{p}_t \in [0,1]$ is termed as \textit{confidence} of the prediction $\hat{y}_t = \argmax \bp_t$. The \textit{accuracy} of a confidence value $p \in [0,1]$, denoted by $\acc(p)$ is the action classification accuracy of frames with maximum probability prediction $\hat{p}_t$ equal to $p$. Ideally, one would like the \acc{} to be high for high values of confidence and vice-versa. A model is calibrated if $\acc(p) = p, \,\forall p\in [0,1]$ and it is called over-confident (or under-confident) if $\acc(p)\! \le\!p$ (or $\acc(p)\!>\!p$). 
The above definition of \acc{} is often made practical by calculating the accuracy for a range of confidence values $\calP \subset [0,1]$, rather than one particular value $p$. Thus the modified definition becomes -
\begin{gather*}
    \acc(\calP) := \frac{\sum_t \bbI\sqBK{\hat{y_t} = y_t}\cdot \bbI\sqBK{\hat{p}_t \in \calP}}{\sum_t \bbI\sqBK{\hat{p}_t \in \calP}}.
\end{gather*}
For $\calP = [0,1]$, the definition of accuracy reduces to the standard classification accuracy. We use this notion of confidence and accuracy to later measure the calibration performance of temporal segmentation in sub-section \ref{subsec:uncertainty_quantification}.

\section{Experiments}\label{sec:results}

\subsection{Datasets, Evaluation, and Implementation:} 
We evaluate on three standard action segmentation benchmarks: Breakfast Actions\cite{kuehne2014language}, 50Salads\cite{stein2013combining} and GTEA\cite{gtea-fathi2011learning}.
\textbf{Breakfast Actions} is a third-person view dataset of 1.7k videos of 52 subjects performing ten high-level tasks for making breakfast.  
On average, the videos are $2.3$ minutes long with $6$ subactions (a total of 48 possible actions).
\textbf{50Salads} has top-view videos of $25$ people preparing $2$ mixed salads each, totally 50 videos with $19$ different sub-actions. 
The videos have average length of $6.4$ minutes and an average of $20$ actions. 
\textbf{GTEA} captures $28$ egocentric videos of $7$ complex activities 
with $11$ sub-actions. The average duration of videos is $1.5$ minutes with $20$ sub-action instances. 

For \textbf{evaluation}, we report Mean-over-frames(MoF), segment-wise edit distance (Edit) and $F1$-scores with IoU thresholds of $0.10$, $0.25$ and $0.50$($F1@\{10, 25, 50\}$). For all three datasets, we use features pre-extracted from an I3D model~\cite{carreira2017quo} pre-trained on Kinetics, and follow the $k$-fold cross-validation averaging to report our final results. Here $k=\{4, 5, 4\}$ for Breakfast, 50Salads and GTEA respectively. 
The evaluation metrics and features follow the convention of other recent temporal video segmentation methods~\cite{sener2020temporal, li2020ms, wang2020boundary, ishikawa2021alleviating}. 

\begin{table*}[t]
\begin{center}
\small{
\begin{tabular}{p{2.7cm} | p{0.5cm}p{0.5cm}p{0.5cm}p{0.5cm}p{0.6cm} | p{0.5cm}p{0.5cm}p{0.5cm}p{0.5cm}p{0.6cm} | p{0.5cm}p{0.5cm}p{0.5cm}p{0.5cm}p{0.6cm}}
\hline
& \multicolumn{5}{c|}{Breakfast} & \multicolumn{5}{c|}{50Salads} & \multicolumn{5}{c}{GTEA} \\
\hline\hline
\textbf{Method} & \multicolumn{3}{c}{$F1@\{10,25,50\}$} & Edit & MoF & \multicolumn{3}{c}{$F1@\{10,25,50\}$} & Edit & MoF & \multicolumn{3}{c}{$F1@\{10,25,50\}$} & Edit & MoF\\
\hline
Base Model $\enc,\bottle,\dec$ & $56.6$ & $52.5$ & $43.4$ & $57.4$ & $65.8$ & 
                $67.5$ & $64.3$ & $53.9$ & $59.1$ & $77.5$ & 
                $87.1$ & $82.6$ & $69.3$ & $81.4$ & $77.3$\\ 
 \textbf{(+)} C2F Ensemble & $64.5$ & $60.4$ & $49.1$ & $63.1$ & $70.2$
                            &$72.3$ & $68.8$ & $57.8$ & $66.6$&$78.4$
                            &$88.1$ & $86.8$ &$73.7$ &$84.1$& $78.5$ \\
 \textbf{(+)} Train Augment & $69.4$ & $65.9$ & $55.1$ & $66.5$ & $73.4$ & 
                 $75.8$ & $73.1$ & $62.3$ & $68.8$ & $79.4$ & 
                 $90.1$ & $87.8$ & $74.9$ & $86.7$ & $79.5$ \\
 \textbf{(+)} Action Loss & $70.1$ & $66.6$ & $56.2$ & $68.2$ & $73.5$ & 
                $76.6$ & $73.0$ & $62.5$ & $69.2$ & $80.1$ & 
                $\mathbf{90.5}$ & $88.5$ & $77.1$ & $\mathbf{87.3}$ & $80.3$ \\
 \textbf{(+)} Test Aug. (\textbf{final}) & $\mathbf{72.2}$ & $\mathbf{68.7}$ & $\mathbf{57.6}$ & $\mathbf{69.6}$ & $\mathbf{76.0}$ 
                & $\mathbf{84.3}$ & $\mathbf{81.8}$ & $\mathbf{72.6}$ & $\mathbf{76.4}$ & $\mathbf{84.9}$ & 
                $90.3$ & $\mathbf{88.8}$ & $\mathbf{77.7}$ & $86.4$ & $\mathbf{80.8}$ \\
 \textbf{(--)} TPP layer $\bottle$ & $69.9$ & $66.6$ & $56.5$ & $66.9$ & $75.1$ &
                $81.7$ & $79.9$ & $71.0$ & $74.0$ & $83.9$ & 
                $89.6$ & $88.3$ & $77.4$ & $86.3$ & $80.4$ \\
\hline\hline
\end{tabular}
}
\end{center}
\caption{\textbf{Ablation study} on each component of our proposal. We gradually add \textbf{(+)} each part of our proposed method to show its effectiveness. To highlight the fact that temporal pyramid pooling is most effective when inputs are of varying resolution, we show its ablation as removal \textbf{(--)} only after we add train and test augmentation to our method stack.}\label{tab:ablation}
\end{table*}

\begin{table*}[t]
\begin{center}
\small{
\begin{tabular}{p{2.3cm} | p{0.5cm}p{0.5cm}p{0.5cm}p{0.6cm}p{0.6cm} | p{0.5cm}p{0.5cm}p{0.5cm}p{0.6cm}p{0.6cm} | p{0.5cm}p{0.5cm}p{0.5cm}p{0.6cm}p{0.6cm}}
\hline
& \multicolumn{5}{c|}{Breakfast} & \multicolumn{5}{c|}{50Salads} & \multicolumn{5}{c}{GTEA}\\
\hline
\textbf{Method} & \multicolumn{3}{c}{$F1@\{10,25,50\}$} & Edit & MoF & \multicolumn{3}{c}{$F1@\{10,25,50\}$} & Edit & MoF & \multicolumn{3}{c}{$F1@\{10,25,50\}$} & Edit & MoF\\
\hline\hline
ED-TCN\cite{TED-lea2017temporal} & -- & -- & -- & -- & $43.3$ & $68.0$ & $63.9$ & $52.6$ & $52.6$ & $64.7$ & $72.2$ & $69.3$ & $56.0$ & -- & $64.0$\\
TDRN\cite{TEDresi-lei2018temporal} & -- & -- & -- & -- & -- & $72.9$ & $68.5$ & $57.2$ & $66.0$ & $68.1$ & $79.2$ & $74.4$ & $62.7$  & $74.1$  & $70.1$ \\
MSTCN\cite{farha2019ms} & $52.6$ & $48.1$ & $37.9$ & $61.7$ & $66.3$ & $76.3$ & $74.0$ & $64.5$ & $67.9$ & $80.7$  & $85.8$ & $83.4$ & $69.8$ & $79.0$ & $76.3$\\
GTRM\cite{huang2020improving} & $57.5$ & $54.0$ & $43.3$ & $58.7$ & $65.0$ & $75.4$ & $72.8$ & $63.9$ & $67.5$ & $82.6$  & -- & -- & -- & -- & -- \\

MSTCN++\cite{li2020ms} & $64.1$ & $58.6$ & $45.9$ & $65.6$ & $67.6$ & $80.7$ & $78.5$ & $70.1$ & $74.3$ & $83.7$ & $87.8$ & $86.2$ & $74.4$ & $82.6$ & $78.9$\\
GatedR\cite{wang2020gated} & $71.1$ & $65.7$ & $53.6$ & $\mathbf{70.6}$ & $67.7$ & $78.0$ & $76.2$ & $67.0$ & $71.4$ & $80.7$ & $89.1$ & $87.5$ & $72.8$ & $83.5$ & $76.7$ \\
BCN\cite{wang2020boundary} & $68.7$	& $65.5$ & $55.0$ & $66.2$ & $70.4$ & $82.3$ & $81.3$ & $\mathbf{74.0}$ & $74.3$ & $84.4$ & $88.5$ & $87.1$ & $77.3$ &	$84.4$ & $79.8$\\
\hline
\textbf{Ours proposed} & $\mathbf{72.2}$ & $\mathbf{68.7}$ & $\mathbf{57.6}$ & $69.6$ & $\mathbf{76.0}$ & $\mathbf{84.3}$ & $\mathbf{81.8}$ & $72.6$ & $\mathbf{76.4}$ & $\mathbf{84.9}$ & $\mathbf{90.3}$ & $\mathbf{88.8}$ & $\mathbf{77.7}$ & $\mathbf{86.4}$ & $\mathbf{80.8}$ \\
\hline\hline
\end{tabular}
}
\end{center}
\caption{\textbf{Comparison with recent related work}. Our proposed model exceeds in most of the scores across all datasets.}\label{tab:final result}
\end{table*}

\textbf{Implementation Details} In our experiments we train using an Adam optimizer for $600$ epochs, with learning rates of $\{10^{-4}, 3\!\times\! 10^{-4}, 5\!\times\!10^{-4}\}$, weight decay of $\{3\times10^{-3}, 10^{-3}, 3\times10^{-4}\}$, batch size of $\{100, 25, 11\}$ and a base window for sampling $w_0$ of $\{10, 20, 4\}$ for Breakfast, 50Salads and GTEA respectively. While choosing the base window we ensure it is small enough not to drop any sub-action or fragments. The ensemble weights used were $\alpha_i=\frac{1}{4} ,\;\forall i\!\geq\!3$ for all three datasets. We find for $i\!\le\!2$, the results of the ensemble degrade due to very coarse temporal resolutions.

\subsection{Ablation Studies \& Model Analysis:} 

In Table~\ref{tab:ablation} we perform a component-wise analysis of our core contributions.  In the first row, we start with the base model $M\!:=\! (\enc\!:\!\bottle\!:\!\dec)$ with the probability outputs $\bp^{(6)}$ from the last decoder layer.  This base model already outperforms other temporal encoder-decoder architectures such as ED-TCN \cite{TED-lea2017temporal} and Residual Deformable ED-TCN \cite{TEDresi-lei2018temporal} (see Table~\ref{tab:final result}).  The gradual addition of each component which we detail below, shows a steady improvement of results.

\textbf{Impact of Coarse-to-Fine Ensemble:} Adding the decoder ensemble to the base model gives a significant improvement in Edit and F1 scores on all three datasets. The Edit score improves by +5.7\%, +7.5 \%, +2.7 \% on Breakfast, 50Salads and GTEA; the extent of improvement directly aligns with the average length of the videos. The improvement over the base model in F1@50 is also a significant +5.7 \%, +3.9 \%, +4.4 \%; this demonstrates the usefulness of our decoder ensembling. With only the ensembling, we can exceed the performance of MS-TCN++ (see Table \ref{tab:final result} row 5) in all the scores on Breakfast Actions dataset.

\textbf{Analysis of Coarse-to-Fine Ensemble:} 
Figure~\ref{fig:layer-analysis} shows a qualitative and quantitative analysis of the ensemble compared to the individual decoder layers $\dec^{(i)}$ which form the ensemble. The left plot shows our model's segmentation output on a sample video where each color denotes a sub-action. The output from the last decoder layer $\dec^{(6)}$ becomes over-fragmented (blue patch), which does not corrupt the ensembled result. Plotting the MoF, Edit, and $F1@50$ scores (see Fig.~\ref{fig:layer-analysis} right), we find that the fourth decoder layer($\dec^{(4)}$) has the strongest individual performance and that these results erode with further decoding.  However, incorporating all the layers in the C2F ensemble yields a significant boost, especially in Edit distance, which is the measure most affected by fragmentation.

\textbf{Number of Layers used in C2F Ensemble:}
We mention that we use ensemble weights with $\alpha_i= \frac{1}{4}, \forall i \geq 3$ and $\alpha_i = 0, \forall i \le 2$. This is found using experimental validation shown in Table \ref{tab:weights_experiment}. The first column shows the un-normalized weights. Here we perform an ablation study in Breakfast dataset on how many layers of Decoder's output are useful for ensembling. We use our final proposed model with C2F Ensemble, Action Loss, Train, and Test Augmentation and only modify the weights $\alpha_i$ in the ensemble to perform the ablation. We see that we get the maximum of Edit, MoF and F1 scores at \textit{Last four} layers, \ie when $\alpha_i=1/4, \forall i \geq 3$ and $\alpha_i=0, i \leq 2$.

\begin{table}[]
    \centering
    \begin{tabular}{|p{3.3cm}|p{0.5cm}p{0.5cm}p{0.5cm}p{0.5cm}p{0.6cm}|}
    \hline
         $\{\alpha_1, \alpha_2, \alpha_3, \alpha_4, \alpha_5, \alpha_6\}$ & \multicolumn{3}{c}{F1@\{10, 25, 50\}} & Edit & MoF \\
         \hline\hline
         $\{0, 0, 0, 0, 0, 1\}$ & $65.7$ & $62.2$ & $52.7$ & $65.3$ & $75.8$\\\hline
         $\{0, 0, 0, 0, 1, 1\}$ & $67.0$ & $63.2$ & $52.9$ & $65.0$ & $75.5$\\\hline
         $\{0, 0, 0, 1, 1, 1\}$ & $67.6$ & $64.2$ & $54.6$ & $66.9$ & $75.3$\\\hline
         $\{0, 0, 1, 1, 1, 1\}$ & $\mathbf{72.2}$ & $\mathbf{68.7}$ & $\mathbf{57.6}$ & $\mathbf{69.6}$ & $\mathbf{76.0}$\\\hline
         $\{0, 1, 1, 1, 1, 1\}$ & $71.4$ & $67.8$ & $57.1$ & $68.8$ & $74.8$\\\hline
         $\{1, 1, 1, 1, 1, 1\}$ & $70.6$ & $66.5$ & $55.6$ & $67.6$ & $74.4$ \\\hline
    \end{tabular}
    \caption{Ablation study on number of layers used in C2F Ensemble on Breakfast dataset.}
    \label{tab:weights_experiment}
\end{table}

\begin{figure*}
\begin{center}
\includegraphics[width=0.9\linewidth]{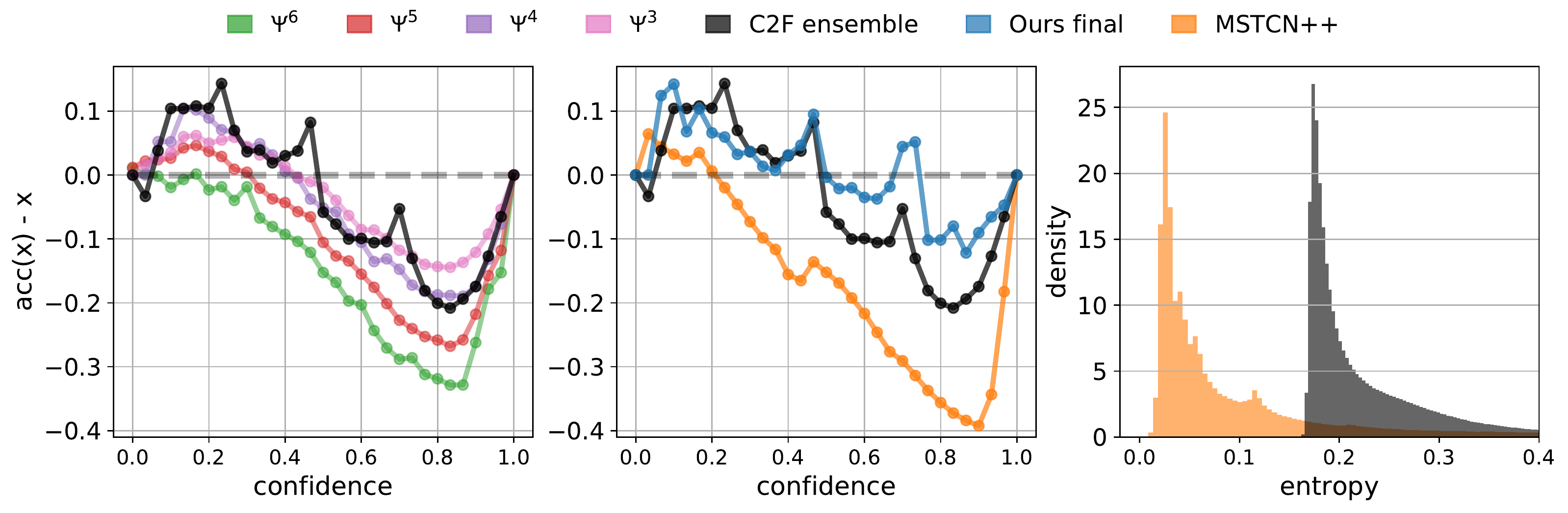}
\end{center}
\caption{\textbf{Uncertainty quantification}: \textit{(Left and middle)} plot show calibration curves of different methods. Unlike the standard calibration curve, we plot the difference between accuracy and confidence in our y-axis. The ideal line is $y\!=\!0$, with $y\!>\!0$ and $y\!<\!0$ denoting under-confidence and over-confidence respectively. The first plot shows the comparison between layers, and the middle plot compares the calibration of MSTCN++, our C2F ensemble, and our final prediction (with test time augment). Performance-wise our final predictions are more calibrated than C2F ensemble, which is more calibrated than any other decoder layer and MSTCN++. \textit{(Rightmost)} plot is the density of the entropy of probability for incorrect predictions. Our C2F ensemble is more uncertain about wrong predictions than MSTCN++.}
\label{fig:uncertainty}
\end{figure*}

\textbf{Impact of Multi-Resolution Feature-Augmentation Strategy:} Table \ref{tab:ablation} row $3$ shows that adding feature augmentation during training provides a boost in all scores across all datasets.
The maximum improvement of scores with train augmentation strategy is seen in Breakfast Action dataset with +3.2\% Mof and +3.4\% Edit scores. In row $5$, we observe additional improvement with test-time augmentation strategy, with the highest increase for 50Salads with +10.1\% F1@50. Interestingly, we also observe some decrements for GTEA; we speculate that this is due to some very short segments getting vanished in some of the coarser windows of test time augmentation. 

\textbf{Impact of Video-Level Action Loss $\calL_{\text{AL}}$:} Table \ref{tab:ablation} row $4$ shows that adding this loss is also beneficial.  
Edit Scores' improvement is +1.7\%, +0.4\%, +0.6\% in Breakfast, 50Salads, and GTEA datasets. The maximum improvement in out-of-activity prediction is on Breakfast Actions because it has the most sub-actions $48$ compared to $19$ and $11$ for 50Salads and GTEA.

\textbf{Temporal Pyramid Pooling layer \bottle:} In the last row of Table \ref{tab:ablation} we show that removing the Temporal Pooling Layer lowers the scores in all the datasets and metrics.  It highlights the importance of including a multi-resolution hidden feature representation at the bottleneck.

\subsection{Comparison with \textit{SOTA}:}\label{subsec:sota_comparison}
Table \ref{tab:final result} compares our performance against recent and related \textit{SOTA} segmentation approaches. All the listed works use the same $I3D$ features and evaluation splits. There are few other works on temporal segmentation which are not directly comparable \cite{ishikawa2021alleviating, selfsupervised-chen2020action}. SSTDA \cite{selfsupervised-chen2020action} uses self-supervised task's adapted features to train the segmentation architecture, Alleviating-Over-segmentation \cite{ishikawa2021alleviating} uses features extracted from fully-trained MS-TCN\cite{farha2019ms} architecture to train segmentation architecture. 

Our model outperforms the \textit{SOTA} scores by +5.6\%, +2.6 \%, +3.2 \% on MoF, F1@50 and F1@25 scores on Breakfast, the largest of three datasets. Our Edit score is slightly lower than 
GateR\cite{wang2020gated}. However, GateR's MoF is -8.3 \% less than ours.  For the smaller 50Salads, we outperform the \textit{SOTA} scores by +2.1 \%, +2.0 \% on Edit and F1@10.  For F1@50, we are slightly lower than BCN\cite{wang2020boundary}, but for all other measures and datasets, BCN is worse.
On GTEA, we outperform SOTA scores on all metrics with +1.0\%, 2.0\%, 1.7\% on MoF, Edit, and F1@25 scores.  We conclude from these strong evaluation scores that our method can generalize to different types of videos and datasets. 

\textbf{Impact of Video Length:} For a closer look, we split the videos into three length categories and tally our results.  In comparison, we train an MS-TCN++\cite{li2020ms} model, which achieves comparable or higher scores than reported in the original paper in all metrics. In the Table \ref{tab:length_exp} we show the MoF $\%$ for various video lengths.  We observe that after training augmentation (row 3), our performance improves regardless of the videos' length. For the longer videos ($\geq 2.5$ mins), our final proposal achieves +5.7\% MoF over the MS-TCN++ model.

\subsection{Evaluation of Complex Activity Recognition:} In Table \ref{tab:high-level-action-results} we show our results on Action Recognition task.  We compare against several methods with varying architectures~\cite{highlevel-girdhar2017actionvlad, highlevel-hussein2019videograph,highlevel-hussein2019timeception, highlevel-hussein2020pic} ranging from graphs to multi-scale self-attention dedicated in their design for long-video recognition. Unlike these works, we adapt a segmentation architecture but find that we can still outperform these other works. For a fair comparison, we use the same splits provided by \cite{highlevel-hussein2019videograph, highlevel-hussein2019timeception, highlevel-hussein2020pic}, which has $\{1357:335\}$ videos for training and evaluation. We use features from I3D pre-trained on Kinetics \cite{carreira2017quo} dataset, and it is \textit{not} fine-tuned for the Breakfast Action dataset. Our base model is +15.4 \% above other methods without fine-tuned features and competitive with other methods that do use fine-tuned features. Adding the subsequent components steadily improves our results surpassing \textit{SOTA} that uses fine-tuned $I3D$ features by +5\% even though our features are not fine-tuned.

\begin{table}
\begin{center}
\small{
\begin{tabular}{l| c c c} 
\hline
\multirow{2}{*}{Method} & \multirow{2}{*}{Feature} & Not Fine & Fine\\
& &  Tuned & Tuned\\
\hline
\hline
ActionVLAD \cite{highlevel-girdhar2017actionvlad} & I3D & $65.5$ & $82.7$ \\
VideoGraph \cite{highlevel-hussein2019videograph} & I3D & $69.5$ & - \\
Timeception \cite{highlevel-hussein2019timeception} & I3D & $71.3$ & $86.9$ \\
Generative \cite{highlevel-kuehne2016end} & FV \cite{IDT-wang2013action} & $73.3$ & - \\
PIC \cite{highlevel-hussein2020pic} & I3D & - & $89.9$ \\
Actor-Focus \cite{highlevel-ballan2021long} & I3D & $72.0$ & $89.9$ \\
\hline
Ours Base Model & I3D & $88.7$ & -\\
\textbf{(+)} C2F Ensemble & I3D & $90.7$ & - \\
\textbf{(+)} Train Aug. & I3D & $92.1$ & - \\
\textbf{(+)} Test Aug. & I3D & $\mathbf{94.9}$ & - \\
\hline\hline
\end{tabular}
}
\end{center}
\caption{\textbf{Complex Activity Recognition:} we outperform previous \textit{SOTA} without using fine-tuned features}
\label{tab:high-level-action-results}
\end{table}

\subsection{Uncertainty Quantification:}\label{subsec:uncertainty_quantification}

As motivated in section \ref{sec:introduction} it is crucial to evaluate models dealing with videos based on uncertainty quantification.
Hence, we evaluate our model by comparing its calibration and prediction uncertainty with the trained MS-TCN++ model of \ref{subsec:sota_comparison}. We follow the notations defined in sub-section \ref{subsec:calibration_notation}. To make the calibration plot, we partition the unit interval $[0,1]$ into $N$ equal length bins $\calP_n := (\frac{n}{N}, \frac{n+1}{N}]$, and compute $\acc_n := \acc(\calP_n)$. We then obtain the 2-D curve $\{(\frac{2n+1}{2N}, \acc_n): 1\!\le\!n\!\le\!N\}$ in which the x-axis denotes confidence (mid-point of interval $\calP_m$) and y-axis denotes its corresponding accuracy. For better visualization, we plot the \textit{difference} between accuracy and confidence in the y-axis. In the first two plots of figure \ref{fig:uncertainty} we compare the calibration performance of predictions from our final proposed stack (with train and test augment), C2F ensemble prediction, MSTCN++, and prediction from different decoder layers. Our final proposal gives the most calibrated result. In addition, our C2F ensemble is more calibrated than MSTCN++. For the last plot, we calculate the \textit{Shannon Entropy} of probability predictions for incorrect segmentation. Higher entropy indicates more uncertainty in prediction. We plot the density of the calculated entropy for MSTCN++ and our C2F ensemble. Our ensemble predictions are more uncertain when the model is wrong.

\begin{figure*}[ht!]
\centering
\includegraphics[width=\linewidth]{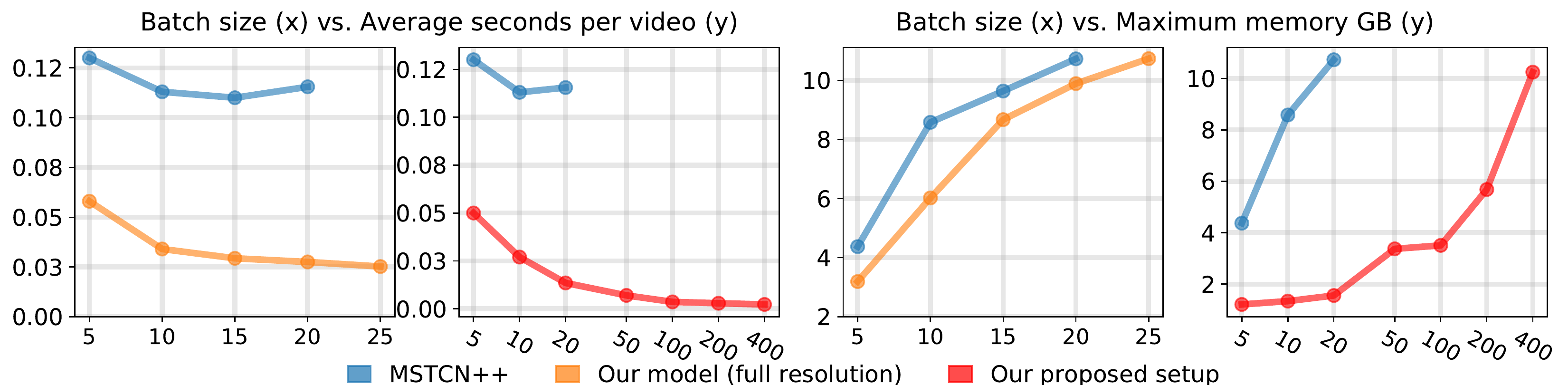}
\caption{\textbf{Computational resources:} In the first two figure, we compare the average time (in seconds) per example used during training on Breakfast dataset. In addition to comparing MSTCN++ with our proposed method (denoted as \textit{our proposed setup}), we also compare the resource used when our proposed method is run with full resolution inputs (denoted by \textit{our model (full resolution)}) as done in MSTCN++. All the graphs terminate when the batch size could not be fit into a single GPU. For better visualization, some graphs use a logarithmic scale on the x-axis (batch size). Our method takes considerably less computing time and memory.}
\label{fig:resources}
\end{figure*}

\begin{table}[]
\begin{center}
\small{
\begin{tabular}{|l|c|c|c|c|}
    \hline
    Duration & $\leq\!1$ min & $>\!1$ and $\le\!2.5$ & $>\!2.5$ min \\
    \hline
    No. of Videos & $534$ & $584$ & $594$ \\
    \hline
    MS-TCN++\cite{li2020ms} & $68.7$ & $70.5$ & $70.2$\\
    Ours C2F Ensemble & $68.9$ & $69.8$ & $69.7$ \\
    \textbf{(+)} Train-Aug & $72.9$ & $72.9$ & $72.7$ \\
    \textbf{(+)} Test-Aug (final) & $\mathbf{73.0}$ & $\mathbf{73.3}$ & $\mathbf{75.9}$ \\
    \hline\hline
\end{tabular}
}
\end{center}
\caption{MoF for varying lengths of videos from Breakfast Dataset.}
\label{tab:length_exp}
\end{table}

\subsection{Computational efficiency:}

In this section, we discuss the computational benefits in terms of memory-usage and compute-time of our proposed method compared to MSTCN++\cite{li2020ms}. 

The computation gains stem from two main reasons:
\begin{enumerate}
    \item \textit{Shrink-then-stretch} of temporal resolution 
    compared to maintaining the same temporal resolution architecture of  MS-TCN/MS-TCN++ \cite{farha2019ms,li2020ms}.
    \item \textit{Feature sub-sampling} obtained from temporal pooling as described in Multi-Resolution Feature Augmentation subsection \ref{subsec:augmentation} compared to using full resolution features as motivated in \cite{farha2019ms, li2020ms}.
\end{enumerate}

We report two metrics: (1) \textit{Average Training Compute Time} per video, (2) \textit{Maximum GPU Memory} usage during training. To report the compute time, we exclude the time taken for data preparation and transfer time between devices, thus only capturing the time taken for the forward and backward pass during training. 
We report all our metrics based on a single \textit{Nvidia GeForce RTX 2080 Ti} GPU with 10.76 Gb of memory. We compare our architecture to the base model MS-TCN++ \cite{li2020ms}.  We further note that other methods in the \textit{SOTA} comparison, like BCN \cite{wang2020boundary} and GatedR \cite{wang2020gated}, all use MS-TCN++ as their base architecture, with additional model components to resolve over-segmentation.  We, therefore, assume that these works would have similar or higher time and memory consumption compared to MS-TCN++. 

In Figure~\ref{fig:resources}, we compare our method with MSTCN++ in terms of the two metrics defined above. MS-TCN++ uses a full-resolution setup, and we show that our model \modelname works best with a sub-sampled version. For a fair comparison, we show our \modelname using a full-resolution set of features (\ie, where the window of temporal pooling is $w_0=1$), similar to MS-TCN++, and also compare \modelname with sub-sampling by a factor of 10, (\ie, where $w_0=10$ as in our final proposal). The blue curve denotes MSTCN++; the orange curve denotes our \modelname with full-resolution features; the red curve denotes our \modelname with sub-sampled features. The x-axis of each plot shows the batch size. 

We 
increase the batch size from 5 to the maximum that fits in the one single RTX2080 GPU (20 for MS-TCN++, 25 for our model at full resolution, and 400 for sub-sampled features).  
In a similar setup, with the maximum batch size, the minimum processing time for a single video for MS-TCN++ is $0.11$ seconds, while our model is $0.03$ seconds at full-resolution and $0.003$ seconds when subsampled, \ie a speedup of more than 30X. In conclusion, we show that our method takes less resources even when we use full resolution features; using sub-sampled features as proposed in our setup allows for even further reduction.  

\section{Details of Encoder-Decoder Architecture}

Here we give the detailed model architecture explained in subsection \ref{subsec:model_architec}. To define our model, we first define a block called \textit{double\_conv} block where \textit{double\_conv(in\_c, out\_c)} = \textit{Conv1D(in\_c, out\_c, kernel=3, pad=1)} $\xrightarrow[]{}$ \textit{BatchNorm1D(out\_c)} $\xrightarrow[]{}$ \textit{ReLU()} $\xrightarrow[]{}$ \textit{Conv1D(out\_c, out\_c, kernel=3, pad=1)} $\xrightarrow[]{}$ \textit{BatchNorm1D(out\_c)} $\xrightarrow[]{}$ \textit{ReLU()}; \textit{in\_c} denotes input channel's dimension and \textit{out\_c} denotes the output channel's dimension. Using this block, we define our model $M$ detailed in the Table \ref{tab:model_arch}. The output from $\dec_i$ is then projected to \textit{number of classes} and followed by a softmax operation to produce probability vectors $\bp^{(i)}$ as described in subsection \ref{subsec:model_architec}. Our model has a total of $4.08$ million trainable parameters.

\begin{table}[t]
\begin{center}
\centering
\small{
\begin{tabular}{|p{0.5cm} | p{1.5cm} | p{3.3cm} | p{1.5cm} |}
\hline
Stage & Input & Model & Output \\\hline\hline

$\enc_0$ & $T_{in} \times 2048$ & \textit{double\_conv(2048, 256)} & $T_{in} \times 256$ \\\hline

\vspace{0.1pt} $\enc_1$ & \vspace{0.1pt} $T_{in} \times 256$ & \vtop{\hbox{\textit{MaxPool1D(2)}}\textit{double\_conv(256, 256)}} & \vspace{0.1pt} $\frac{T_{in}}{2} \times 256$ \\\hline

\vspace{0.1pt} $\enc_2$ & \vspace{0.1pt} $\frac{T_{in}}{2} \times 256$ & \vtop{\hbox{\textit{MaxPool1D(2)}}\textit{double\_conv(256, 256)}} & \vspace{0.1pt} $\frac{T_{in}}{4} \times 256$ \\\hline

\vspace{0.1pt} $\enc_3$ & \vspace{0.1pt} $\frac{T_{in}}{4} \times 256$ & \vtop{\hbox{\textit{MaxPool1D(2)}}\textit{double\_conv(256, 128)}} & \vspace{0.1pt} $\frac{T_{in}}{8} \times 128$ \\\hline

\vspace{0.1pt} $\enc_4$ & \vspace{0.1pt} $\frac{T_{in}}{8} \times 128$ & \vtop{\hbox{\textit{MaxPool1D(2)}}\textit{double\_conv(128, 128)}} & \vspace{0.1pt} $\frac{T_{in}}{16} \times 128$ \\\hline

\vspace{0.1pt} $\enc_5$ & \vspace{0.1pt} $\frac{T_{in}}{16} \times 128$ & \vtop{\hbox{\textit{MaxPool1D(2)}}\textit{double\_conv(128, 128)}} & \vspace{0.1pt} $\frac{T_{in}}{32} \times 128$ \\\hline

\vspace{0.1pt} $\enc_6$ & \vspace{0.1pt} $\frac{T_{in}}{32} \times 128$ & \vtop{\hbox{\textit{MaxPool1D(2)}}\textit{double\_conv(128, 128)}} & \vspace{0.1pt} $\frac{T_{in}}{64} \times 128$ \\\hline

\vspace{2pt} $\bottle$ & \vspace{2pt} $\frac{T_{in}}{64} \times 128$ & \vtop{\hbox{\textit{MaxPool1D(2, 3, 5, 6)}} \textit{conv1d(in\_c=132, out\_c=132, k=3, p=1)}} & \vspace{2pt} $\frac{T_{in}}{64} \times 132$ \\\hline

\vspace{2pt} $\dec_1$ & \vtop{\hbox{$\frac{T_{in}}{64} \times 132$} $\frac{T_{in}}{32} \times 128$} & \vtop{\hbox{\textit{Upsample1D(2)}} \hbox{\textit{concat\_$\enc_5$(132, 128)}} \textit{double\_conv(260, 128)}} & \vspace{2pt} $\frac{T_{in}}{32} \times 128$ \\\hline

\vspace{2pt} $\dec_2$ & \vtop{\hbox{$\frac{T_{in}}{32} \times 128$} $\frac{T_{in}}{16} \times 128$} & \vtop{\hbox{\textit{Upsample1D(2)}} \hbox{\textit{concat\_$\enc_4$(128, 128)}} \textit{double\_conv(256, 128)}} & \vspace{2pt} $\frac{T_{in}}{16} \times 128$ \\\hline

\vspace{2pt} $\dec_3$ & \vtop{\hbox{$\frac{T_{in}}{16} \times 128$} $\frac{T_{in}}{8} \times 128$} & \vtop{\hbox{\textit{Upsample1D(2)}} \hbox{\textit{concat\_$\enc_3$(128, 128)}} \textit{double\_conv(256, 128)}} & \vspace{2pt} $\frac{T_{in}}{8} \times 128$ \\\hline

\vspace{2pt} $\dec_4$ & \vtop{\hbox{$\frac{T_{in}}{8} \times 128$} $\frac{T_{in}}{4} \times 256$} & \vtop{\hbox{\textit{Upsample1D(2)}} \hbox{\textit{concat\_$\enc_2$(128, 256)}} \textit{double\_conv(384, 128)}} & \vspace{2pt} $\frac{T_{in}}{4} \times 128$ \\\hline

\vspace{2pt} $\dec_5$ & \vtop{\hbox{$\frac{T_{in}}{4} \times 128$} $\frac{T_{in}}{2} \times 256$} & \vtop{\hbox{\textit{Upsample1D(2)}} \hbox{\textit{concat\_$\enc_1$(128, 256)}} \textit{double\_conv(384, 128)}} & \vspace{2pt} $\frac{T_{in}}{2} \times 128$ \\\hline

\vspace{2pt} $\dec_6$ & \vtop{\hbox{$\frac{T_{in}}{2} \times 128$} $T_{in} \times 256$} & \vtop{\hbox{\textit{Upsample1D(2)}} \hbox{\textit{concat\_$\enc_0$(128, 256)}} \textit{double\_conv(384, 128)}} & \vspace{2pt} $T_{in} \times 128$ \\\hline
\end{tabular}}
\end{center}
\caption{Encoder-Decoder Architecture $M = (\enc, \bottle, \dec)$}
\label{tab:model_arch}
\end{table}

\section{Conclusion} We design a temporal encoder-decoder model with a coarse-to-fine ensemble of decoding layers which achieves state-of-the-art performance in temporal segmentation.  Our model produces calibrated predictions with better uncertainty measures which is otherwise crucial for real-world deployment.  In addition, we propose a simple and computationally effective augmentation strategy at the feature level which significantly improves results. Such an augmentation strategy can be applied in other works in segmentation or sequence processing. Interestingly, our segmentation architecture allows us to achieve state-of-the-art performance in the complex activity recognition task, opening the  possibility for further investigation along this front.


{\small
\bibliographystyle{ieee_fullname}
\bibliography{temporal}
}

\end{document}